\documentclass[runningheads]{llncs}
\usepackage{times}
\usepackage{xspace}
\usepackage{amsfonts}
\usepackage{amssymb}
\usepackage{mathptm}
\usepackage{latexsym}
\usepackage{calc}
\usepackage{ifthen}
\usepackage{graphicx}
\usepackage{wrapfig}
\usepackage{rotating}
\usepackage{hyperref}

\usepackage{tikz}
\usetikzlibrary{positioning}
\usetikzlibrary{arrows}
\usetikzlibrary{calc}

\graphicspath{{figures/}}

\newcommand{\verifai}{{\sc{VerifAI}}\xspace}
\newcommand{\scenic}{{\sc{Scenic}}\xspace}

\newcommand{\smvertspace}{{\vspace*{-2mm}}}
\newcommand{\medvertspace}{{\vspace*{-5mm}}}

\newcounter{myctr}

\newenvironment{myitemize}{\begin{list}{$\bullet$}
{\setlength{\topsep}{0.3mm}\setlength{\itemsep}{0.25mm}
\setlength{\parsep}{0.1mm}
\setlength{\itemindent}{0mm}\setlength{\partopsep}{0mm}
\setlength{\labelwidth}{15mm}
\setlength{\leftmargin}{3mm}}}{\end{list}}

\begin{document}
\title{\verifai: A Toolkit for the Design and Analysis of Artificial Intelligence-Based Systems\thanks{This work was supported in part by NSF grants 1545126 (VeHICaL), 1646208, 1739816, and 1837132, the DARPA BRASS program under agreement number FA8750-16-C0043, the DARPA Assured Autonomy program, the iCyPhy center, and Berkeley Deep Drive. We acknowledge the support of NVIDIA Corporation via the donation of the Titan Xp GPU used for this research.}}
%
\titlerunning{\verifai}
%
\newcommand{\repeatthanks}{\textsuperscript{\thefootnote}}

\author{Tommaso Dreossi\thanks{Equal contribution.} \and Daniel J. Fremont\repeatthanks \and Shromona Ghosh\repeatthanks
\and Edward Kim \and Hadi Ravanbakhsh \and Marcell Vazquez-Chanlatte
\and Sanjit A. Seshia}
\authorrunning{Dreossi, Fremont, Ghosh, et al.}
%
\institute{University of California, Berkeley, USA}

\maketitle              
\begin{abstract}
We present \verifai, a software toolkit for the formal design and analysis of 
systems that include artificial intelligence (AI) and machine learning (ML)
components. \verifai particularly seeks to address challenges with applying
formal methods to perception and ML components, including those based on 
neural networks, and to model and analyze
system behavior in the presence of environment uncertainty.
We describe the initial version of \verifai which centers on simulation
guided by formal models and specifications. 
Several use cases are illustrated with examples, including temporal-logic falsification,
model-based systematic fuzz testing, parameter synthesis, counterexample analysis,
and data set augmentation.

\end{abstract}
%
%
%
\section{Introduction}
\label{sec:introduction}


The increasing use of artificial intelligence (AI) and
machine learning (ML) in systems, including safety-critical
systems, has brought with it a pressing need for formal methods
and tools for their design and verification.
However, AI/ML-based systems, such as autonomous vehicles,
have certain characteristics that make the application of
formal methods very challenging~\cite{seshia-arxiv16}.
First, several uses of AI/ML are for {\em perception},
the use of computational systems to mimic human perceptual
tasks such as object recognition and classification, 
conversing in natural language, etc.
For such perception components, writing a formal specification
is extremely difficult, if not impossible. Additionally, the
signals processed by such components can be very high-dimensional, 
such as streams of images or LiDAR data.
Second, {\em machine learning} being a dominant paradigm in
AI, formal tools must be compatible with the data-driven
design flow for ML and also be able to handle the 
complex, high-dimensional structures in ML components such as
deep neural networks.
Third, the {\em environments} in which AI/ML-based systems operate
can be very complex, with considerable uncertainty even about how many
(which) agents are in the environment (both human and robotic), 
let alone about their intentions
and behaviors.
As an example, consider the difficulty in
modeling urban traffic environments
in which an autonomous car must operate.
Indeed, AI/ML is often introduced into these systems
precisely to deal with such complexity and uncertainty!
From a formal methods perspective, this makes it very hard to
create realistic environment models with respect to which
one can perform verification or synthesis.

In this paper, we introduce the \verifai toolkit,
our initial attempt to address the three challenges
--- perception, learning, and environments --- that are
outlined above. 
\verifai takes the following approach:
\begin{myitemize}
\item {\em Perception:} 
A perception component maps a concrete feature space (e.g. pixels)
to an output such as a prediction or state estimate. 
To deal with the lack of specification for perception components, 
\verifai analyzes them in the context of a closed-loop system
using a system-level specification.
Moreover, to scale to complex high-dimensional feature spaces,
\verifai operates on an {\em abstract feature space}
(or {\em semantic feature space})~\cite{dreossi-cav18}
that describes semantic aspects of
the environment being perceived, not the raw features such as pixels.

\item {\em Learning:}
\verifai aims to not only analyze the behavior of ML components
but also use formal methods for their (re-)design. To this end, it
provides features to 
(i) design the data set for training and testing~\cite{dreossi-ijcai18},
(ii) analyze counterexamples to gain insight into 
mistakes by the ML model,
as well as 
(iii) synthesize parameters, including hyper-parameters for training
algorithms and ML model parameters.

\item {\em Environment Modeling:}
Since it can be difficult, if not impossible, to exhaustively model
the environments of AI-based systems, \verifai aims to provide ways
to capture a designer's assumptions about the environment, including
distribution assumptions made by ML components, and to 
describe the abstract feature space in an intuitive, declarative manner.
To this end, \verifai provides users with \scenic~\cite{scenic},
a probabilistic programming language for modeling environments.
\scenic, combined with a renderer or simulator for generating sensor
data, can produce semantically-consistent input for perception
components.

\end{myitemize}

\verifai is currently focused on AI-based
cyber-physical systems (CPS), although its basic ideas can
also be applied to other AI-based systems. 
As a pragmatic choice, we focus on simulation-based verification,
where the simulator is treated as a black box, so as to be broadly
applicable to the range of simulators used in industry.
The input to \verifai 
is a ``closed-loop'' CPS model, comprising a composition
of the AI-based system under verification with an environment model,
and a property on the closed-loop model. The AI-based system 
typically comprises
a perception component, a planner/controller, 
and the plant (i.e., the system under control). 
\verifai's output depends on the feature being exercised by
the user. The current version offers the following use cases:
(1) temporal-logic falsification;
(2) model-based fuzz testing;
(3) counterexample-guided data augmentation;
(4) counterexample (error table) analysis;
(5) hyper-parameter synthesis,
and
(6) model parameter synthesis.
To our knowledge, \verifai is the first tool to offer this suite
of use cases in an integrated fashion, unified by a common
representation of an abstract feature space with an accompanying
modeling language and search algorithms
over this feature space.
The problem of temporal-logic falsification or simulation-based
verification of CPS models is well studied and several
tools exist (e.g.~\cite{annpureddyLFS11,duggirala2015c2e2}); 
these techniques have been extended to CPS models with 
ML components by us
and others~\cite{dreossi-nfm17,tuncali-ivs18}.
Work on verification of ML components, especially neural networks 
(e.g.,~\cite{wicker2018feature,gehr2018ai}), is complementary to the system-level analysis
performed by \verifai.
Fuzz testing based on formal models is common in software engineering
(e.g.~\cite{godefroid2008grammar}) but our work is unique in 
the CPS context.
Property-directed parameter synthesis for hybrid systems 
has also been studied 
well in the formal methods/CPS community (e.g.~\cite{donze2009parameter}), 
and we can leverage these results in \verifai.
Finally, to our knowledge, our work on 
augmenting training/test data sets~\cite{dreossi-ijcai18}, 
implemented in \verifai, is the first use
of formal techniques for this purpose.
In Sec.~\ref{sec:structure}, we describe how the tool is structured
so as to provide the above features.
Sec.~\ref{sec:features} illustrates the use cases via examples
from the domain of autonomous driving.

\section{\verifai Structure and Operation}
\label{sec:structure}

\begin{figure}
\centering
\includegraphics[width=0.8\columnwidth]{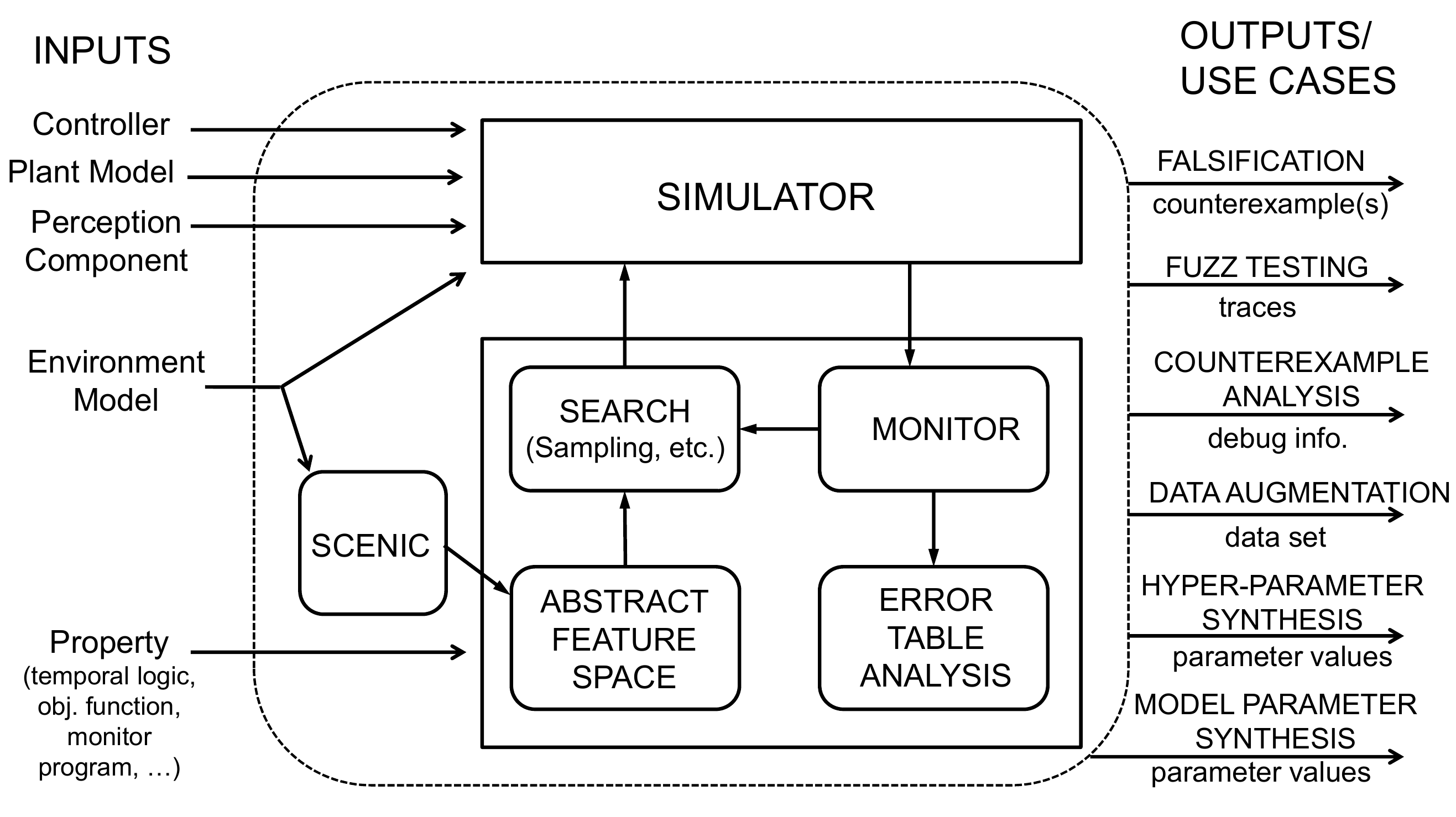}
\caption{Structure and Operation of \verifai.\label{fig:verifai-tool-flow}}
\smvertspace
\end{figure}

The current version of the \verifai toolkit is focused on 
simulation-based analysis and design of 
AI components for perception or control, potentially those using
machine learning, in the context of a closed-loop cyber-physical
system.
Fig.~\ref{fig:verifai-tool-flow} depicts the structure and operation
of the toolkit.

\noindent
{\bf Inputs and Outputs:}
In order to use \verifai, a user must first set up a simulator
for the domain of interest. As we explain in Sec.~\ref{sec:features},
we have experimented with multiple robotics simulators and provide
these in the artifact accompanying the paper. Once this step is done,
the user begins by constructing the inputs to \verifai, including
(i) a simulatable model of the system, including
code for one or more controllers and perception components,
and a dynamical model of the system being controlled (e.g., vehicle);
(ii) a probabilistic model of the environment, specifying constraints on 
the workspace, the locations of agents and objects, and the dynamical behavior of agents, and
(iii) a property on the composition of the system and its environment
(the simulator defines the form of composition).
\verifai is implemented in Python as we found this language to be
the easiest to interoperate with machine learning and AI libraries
and simulators across platforms.
The code for the controller and perception component can be 
arbitrary executable code, typically in a language such as Python or C.
The environment model typically involves two steps: first, in
the simulator, the different agents have to be set up using 
the interface provided by the simulator; then, constraints about
the agents and the workspace can be declaratively specified using
the \scenic probabilistic programming language developed by some of 
the authors~\cite{scenic}.
Finally, the property to be checked can be expressed in multiple
ways depending on the use case of \verifai being exercised,
including metric temporal logic~\cite{alur:mtl,pymtl},
objective functions, and even executable code to monitor a
property.
The output of \verifai depends on the feature being invoked.
For falsification, one or more counterexamples (simulation traces)
are produced showing how the property is violated~\cite{dreossi-nfm17}.
For fuzz testing, one or more traces are produced from the distribution
of behaviors expressed by the environment model described by
the \scenic language~\cite{scenic}.
Error table analysis involves collecting counterexamples generated
by the falsifier into a table, on which we perform analysis to
identify features that are correlated with property failures.
Data augmentation uses falsification and error table analysis to
generate additional data for training and testing an ML component~\cite{dreossi-ijcai18}.
Finally, the property-driven
synthesis of model parameters or hyper-parameters 
generates as output a parameter evaluation that satisfies the
specified property.

\noindent
{\bf Tool Structure:}
\verifai is composed of four main modules, as described below:
\begin{myitemize}
\item {\em Abstract Feature Space and \scenic Modeling Language:}
The abstract feature space is a compact representation of the 
possible configurations of the simulation.
Abstract features can represent parameters of the environment, 
controllers, or of machine learning components.
For example, when analyzing a visual perception system for an 
autonomous car, we might use an abstract feature space consisting of 
the initial poses and types of all vehicles on the road.
Note that this abstract feature space, as compared to the 
concrete feature space of pixels used as input to the controller, 
is better suited to the analysis of the overall closed-loop system 
(e.g. finding conditions under which the car might crash).

\verifai{} provides two ways to construct abstract feature spaces.
They can be constructed hierarchically, starting from basic domains 
such as hyperboxes and finite sets and combining these into 
structures and arrays.
For example, we could define a space for a car as a structure combining 
a 2D box for position with a 1D box for heading, and then create an 
array of these to get a space for several cars.
Alternatively, \verifai{} allows a feature space to be defined using 
a program in the \scenic language~\cite{scenic}.
\scenic{} provides convenient syntax for describing geometric 
configurations and initial conditions for agents, and, as a 
probabilistic programming language, allows placing a distribution over 
the feature space which can be conditioned by declarative constraints.

\item {\em Searching the Feature Space:}
Once the abstract feature space is defined, the next step is to
search that space to find simulations that violate the property or
produce other interesting behaviors. Currently, \verifai uses a 
suite of sampling methods (both active and passive) for this purpose, but in the future we
expect to also integrate directed or exhaustive search methods
including those from the adversarial machine learning literature
(e.g., see~\cite{dreossi-cav18}).
Passive samplers, which do not use any feedback from the simulation,
include uniform random sampling, simulated annealing, and Halton sequence generation~\cite{halton1960efficiency} (a quasi-random deterministic sequence with low-discrepancy guarantees we found effective for falsification~\cite{dreossi-nfm17}).
Distributions defined using \scenic{} are also passive in this sense.
Active samplers, whose selection of samples is informed by feedback from previous simulations, include cross-entropy and Bayesian optimization sampling. The former selects samples and updates the
prior distribution by minimizing cross-entropy; the latter updates the prior from the posterior over a user-provided objective function, e.g.~the satisfaction level of a specification or the loss of an analyzed model.

\item {\em Property Monitor:}
Trajectories generated by the simulator are evaluated by the monitor 
that produces a score for a given property or objective function.
By default, \verifai supports the Metric Temporal Logic~\cite{alur:mtl} 
(MTL) via the\\\texttt{py-metric-temporal-logic}~\cite{pymtl} package.
Both Boolean and quantitative semantics of MTL are supported.
The result of the monitor can be used to output falsifying traces
and also as feedback to the search procedure to direct the sampling
(search) towards falsifying scenarios.

\item {\em Error Table Analysis:}
Counterexamples are stored in a data structure called the error 
table, whose rows are counterexamples and columns are abstract features.
The error table can be used offline to debug (explain) the generated 
counterexamples or online to drive the sampler
towards particular areas of the abstract feature space. 
\verifai provides different techniques for error table analysis 
depending on the end use (e.g., counter-example analysis or 
data set augmentation), including principal component analysis (PCA)
for ordered feature domains and subsets of the most recurrent values
for unordered domains (further details are in~\cite{dreossi-ijcai18}).

\end{myitemize}
The communication between \verifai and the simulator is implemented in a client-server fashion using IPv4 sockets, 
where \verifai sends configurations to the simulator which then
returns trajectories (traces).
This implementation allows easy interfacing to a 
simulator and even with multiple simulators at the same time.


\section{Features and Case Studies}\label{sec:features}

This section illustrates the main features of \verifai{} through case studies demonstrating its various use cases and simulator interfaces.
Specifically, we demonstrate model falsification and fuzz testing of an autonomous car controller, data augmentation and error table analysis for a convolutional neural network, and model and hyperparameter tuning for a reinforcement learning-based controller.
\smvertspace

\subsection{Falsification and Fuzz Testing\label{sec:falsification}}
\smvertspace

\verifai offers a convenient way to debug systems through systematic testing.
Given a model and a specification, the tool can use active sampling to
automatically search for inputs driving the model towards a 
violation of the specification.
\verifai{} can also perform model-based fuzz testing,
exploring random variations of a scenario guided by formal constraints.
To demonstrate falsification and fuzz testing, we consider
two scenarios involving self-driving cars simulated with the robotics simulator Webots~\cite{Webots}. For the experiments reported here, we used Webots 2018 which is commercial software.

In the first example, we falsify the controller of an autonomous vehicle that must navigate around a disabled car and traffic cones which are blocking the road.
The controller is responsible for safely maneuvering around the cones. To achieve this, we implemented a hybrid controller. Initially, the car tries to remain in its lane using a line detector based on standard computer vision (non-ML) techniques\footnote{\url{https://github.com/ndrplz/self-driving-car/blob/master/project\_1\_lane\_finding\_basic/lane\_detection.py}}.
At the same time, a neural network (based on squeezeDet~\cite{squeezedet}) estimates the distance to the cones.
When the distance is less than $15$ meters, the car begins a lane-changing maneuver, switching back to lane-following once the cones are avoided.

The correctness of the autonomous vehicle is characterized by an MTL 
formula requiring the vehicle to maintain a minimum distance from the
traffic cones and avoid overshoot while changing lanes.
The task of the falsifier is to find small perturbations of the initial scene (generated
by \scenic{}) which cause the vehicle to violate this specification.
We allowed perturbations of the 
initial positions and orientations of all objects, the color of the disabled car,
and the cruising speed and reaction time of the ego car.

\begin{figure}[tb]
\centering
\includegraphics[width=0.29\textwidth]{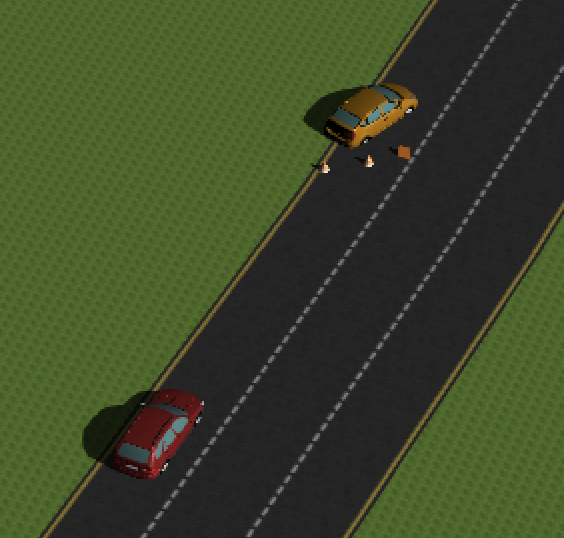}
\hspace{0.02\textwidth}
\includegraphics[width=0.57\textwidth]{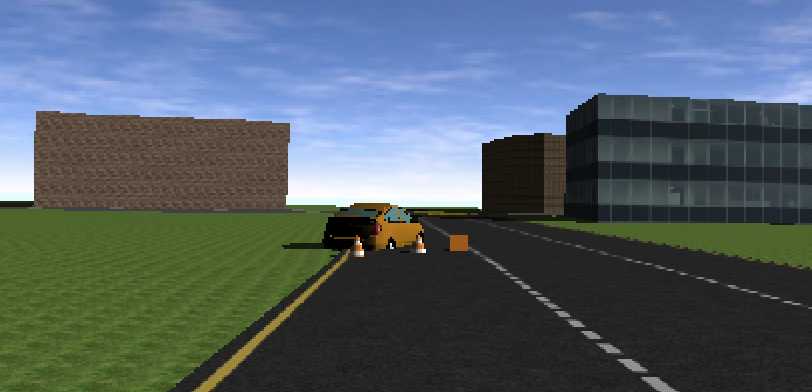}
\caption{A falsifying scene discovered by \verifai{}.
The neural network misclassifies the traffic cones because of the orange vehicle in the background, leading to a crash.
Left: bird's-eye view. Right: dash-cam view, as processed by the neural network.\label{fig:cones_falsif}}
\smvertspace
\end{figure}

Our experiments showed that active samplers driven by the robustness 
of the monitored MTL specification can efficiently discover
scene perturbations that confuse the controller and lead the autonomous vehicle into faulty behavior.
One such counterexample is shown in 
Fig.~\ref{fig:cones_falsif}. The falsifier automatically discovered
that the neural network is not able to detect traffic cones when the car behind them is orange. In this particular case, the
lane change is begun too late and a crash with the disabled vehicle occurs.

\begin{figure}[tb]
\centering
\begin{minipage}{0.45\textwidth}
\includegraphics[width=\textwidth]{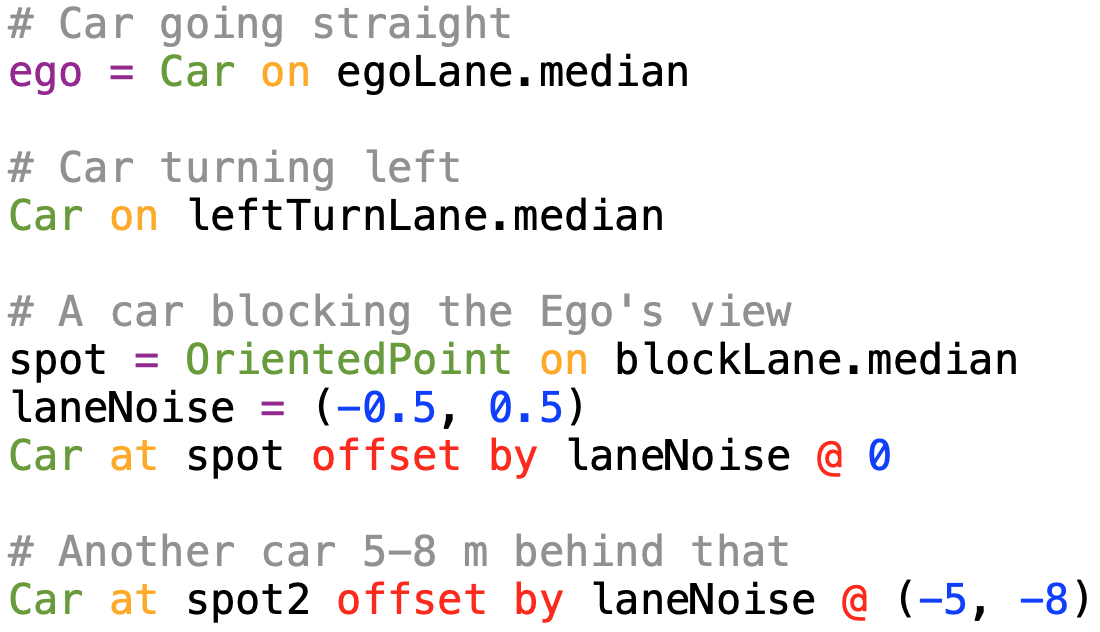}
\end{minipage}
\hspace{0.03\textwidth}
\begin{minipage}{0.5\textwidth}
\includegraphics[width=\textwidth,clip,trim=50 10 175 50]{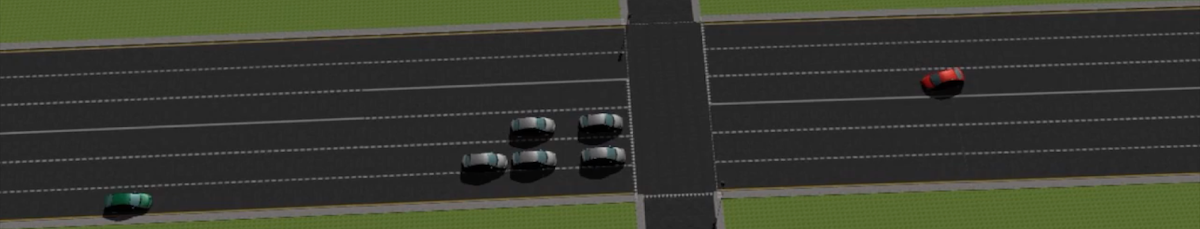}
\includegraphics[width=\textwidth,clip,trim=50 10 175 50]{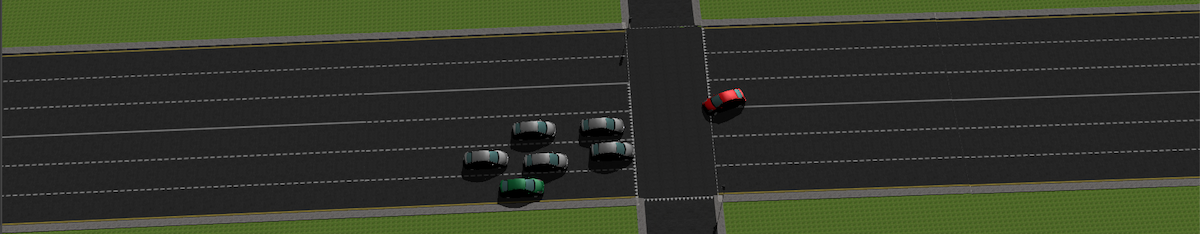}
\includegraphics[width=\textwidth,clip,trim=50 10 175 50]{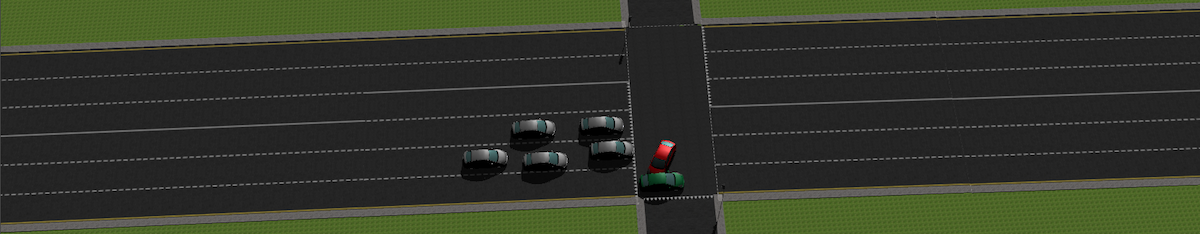}
\end{minipage}
\caption{Left: Partial \scenic{} program for the crash scenario.
\texttt{Car} is an object class defined in the Webots world model (not shown), \texttt{on} is a \scenic{} \emph{specifier} positioning the object uniformly at random in the given region (e.g. the median line of a lane), \texttt{(-0.5, 0.5)} indicates a uniform distribution over that interval, and \texttt{X @ Y} creates a vector with the given coordinates (see \cite{scenic} for a complete description of \scenic{} syntax).
Right: 1) initial scene sampled from the program; 2) the red car begins its turn, unable to see the green car; 3) the resulting collision.\label{fig:accident}}
\smvertspace
\end{figure}

In our second experiment, we used \verifai{} to simulate variations on an actual accident involving an autonomous vehicle.\footnote{March 2017 accident in Tempe, AZ. See \url{https://www.12news.com/article/news/local/valley/self-driving-uber-crashes-in-tempe/425480754}.}
In this accident, an autonomous car is proceeding straight through an intersection when hit by a human turning left.
Neither car was able to see the other until immediately before impact because of two lanes of stopped traffic.
Fig.~\ref{fig:accident} shows a (simplified) \scenic{} program we wrote to reproduce the accident, allowing variation in the initial positions of the cars.
We then ran simulations from random initial conditions sampled from the program, with the turning car using a controller trying to follow the ideal left-turn trajectory computed from OpenStreetMap data using the Intelligent Intersections Toolbox~\cite{guideways}.
The car going straight used a controller which either maintained a constant velocity or began emergency breaking in response to a message from a simulated ``smart intersection'' warning about the turning car.
By sampling variations on the initial conditions, we could determine how much advance notice is necessary for such a system to robustly avoid an accident.

\subsection{Data Augmentation and Error Table Analysis}
\smvertspace

Data augmentation is the process of supplementing training sets with 
the goal of improving the performance of ML models. Typically, 
datasets are augmented with transformed versions of preexisting training examples.
In~\cite{dreossi-ijcai18}, we showed that augmentation with counterexamples is also an effective
method for model improvement.
\begin{wrapfigure}{r}{0.5\textwidth}
\begin{center}
\includegraphics[width=0.48\textwidth]{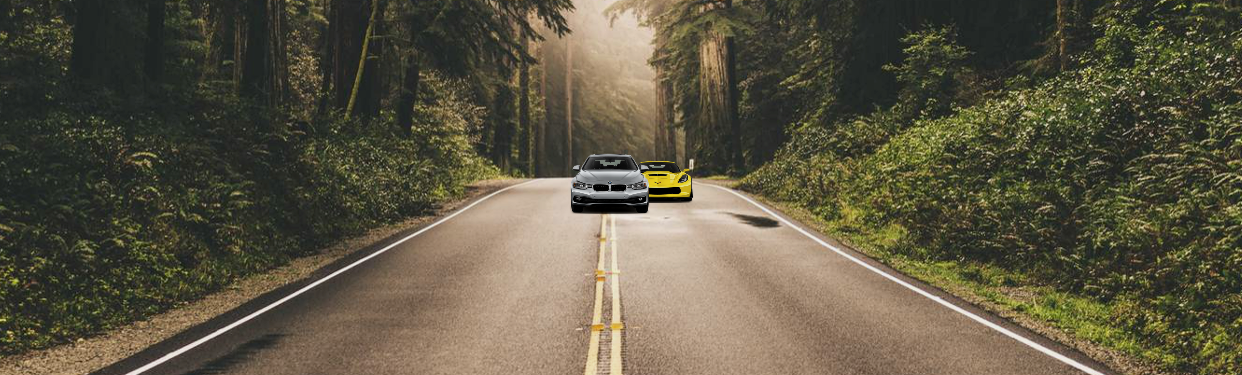}
\hspace{0.02\textwidth}
\end{center}
\vspace{-0.5cm}
\caption{This image generated by our renderer was misclassified by the NN. The network reported detecting only one car when there were two.\label{fig:data_augmentation}}
\smvertspace
\end{wrapfigure}
\verifai implements a counterexample-guided augmentation scheme,
where a falsifier (see Sec.~\ref{sec:falsification}) generates misclassified
data points that are then used to augment the original training set.
The user can choose among different
sampling methods, with passive samplers suited to generating diverse sets of data points while active samplers can efficiently generate similar counterexamples.
In addition to the counterexamples themselves, \verifai also returns an error table aggregating information on the misclassifications that can be
used to drive the retraining process. Fig.~\ref{fig:data_augmentation} shows the rendering of a misclassified sample generated  by our falsifier.

For our experiments, we implemented a renderer of road scenario images and tested the quality of our augmentation scheme on the squeezeDet convolutional neural 
network~\cite{squeezedet}.
We adopted three techniques to select augmentation images: 1) randomly sampling from the error table, 2) selecting the top $k$-closest (similar) samples from the error table, and 3) using PCA analysis to generate new samples.
For details on the image renderer and the results of counterexample-driven augmentation, see~\cite{dreossi-ijcai18}.

\subsection{Model Robustness and Hyperparameter Tuning}
\smvertspace

In this final section, we demonstrate how \verifai can be used
to tune test parameters and hyperparameters of AI systems.
For the following case studies, we use OpenAI Gym~\cite{brockman2016openai},
a framework for experimenting with reinforcement learning algorithms.\footnote{\url{https://github.com/openai/gym}}

\begin{figure}
  \begin{minipage}[c]{0.5\textwidth}
  \centering
    \includegraphics[width=\textwidth]{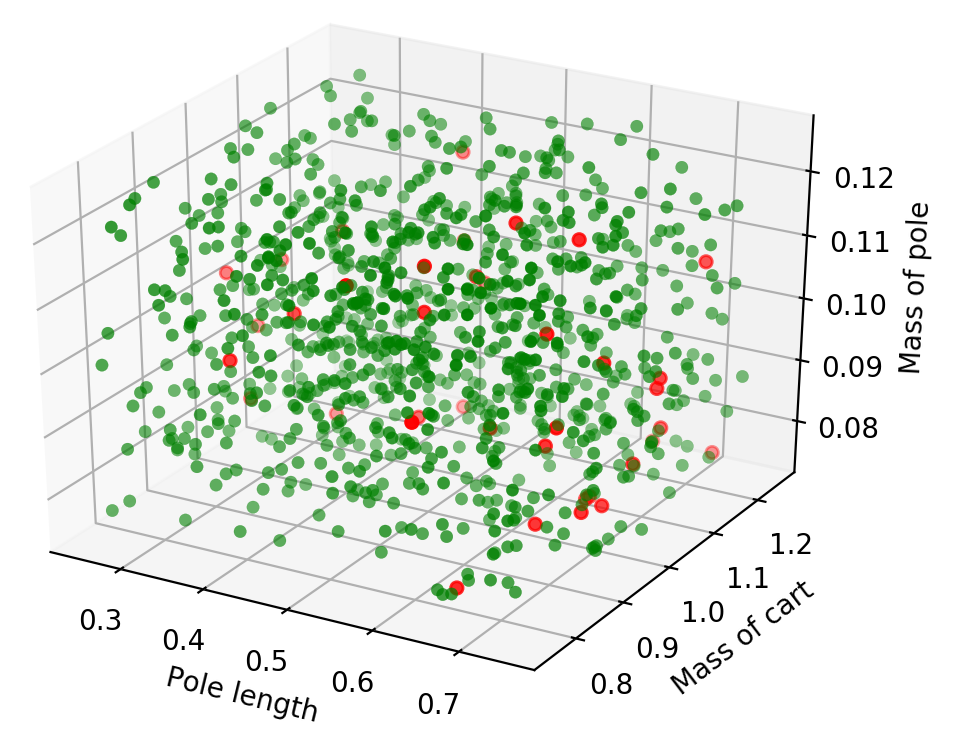}
  \end{minipage}\hfill
  \begin{minipage}[c]{0.45\textwidth}
    \caption{The green dots represent model parameters for which the cart-pole controller behaved correctly, while the red dots indicate specification violations.
    Out of 1000 randomly-sampled model parameters, the controller failed to satisfy the specification 38 times.
    } \label{fig:cartpole_falsif}
  \end{minipage}
  \smvertspace
\end{figure}

First, we consider the problem of testing the robustness
of a learned controller for a cart-pole, i.e., a cart that balances
an inverted pendulum. We trained a neural network to control the cart-pole 
using Proximal Policy Optimization algorithms~\cite{SchulmanWDRK17} with 100k training episodes.
We then used \verifai to test the robustness of the
learned controller, varying the initial lateral position
and rotation of the cart as well as the mass and length of the pole.
Even for apparently robust controllers, \verifai was able to discover 
configurations for which the cart-pole failed to self-balance.  Fig.~\ref{fig:cartpole_falsif} shows 1000 iterations of the falsifier, where sampling was guided by the reward function used by OpenAI to train the controller.
This function provides a negative reward if the cart moves more than 2.4 m or if at any time the angle maintained by the pole is greater than 12 degrees.
For testing, we slightly modified these thresholds.

Finally, we used \verifai to study the effects of hyperparameters when training a neural network controller for a mountain car. In this case, the controller must learn to exploit momentum in order to climb a steep hill. 
Here, rather than searching for counterexamples, we look for a set of hyperparameters under which the network \emph{correctly} learns to control the car.
Specifically, we explored the effects of using different training algorithms (from a discrete set of choices) and the size of the training set.
We used the \verifai{} falsifier to search the hyperparameter space, guided again by the reward function provided by OpenAI Gym (here the distance from the goal position), but negated so that falsification implied finding a controller which successfully climbs the hill.
In this way \verifai{} built a table of safe hyperparameters.
This table can be further analyzed to find the hyperparameters which most affect the training process and which hyperparameters the training is most robust to. This can be done studying the variation across the parameters using PCA analysis.


\section{Conclusion}
\label{sec:conclusion}
\medvertspace
We presented \verifai, a toolkit for the formal design and
analysis of AI/ML-based systems. Although our focus has been
on CPS, we note that \verifai's architecture should be
applicable to simulation-based analysis of other systems.
We also plan to extend the toolkit beyond directed simulation
to include symbolic techniques and incorporate synthesis methods
(e.g.~\cite{alshiekh-aaai18}).
The artifact accompanying this paper includes all the examples
illustrating the various features of \verifai described in
Sec.~\ref{sec:features}, with detailed instructions and
expected output.

%
%
\bibliographystyle{splncs04}
\bibliography{biblio}

\begin{thebibliography}{10}
\providecommand{\url}[1]{\texttt{#1}}
\providecommand{\urlprefix}{URL }
\providecommand{\doi}[1]{https://doi.org/#1}

\bibitem{alshiekh-aaai18}
Alshiekh, M., Bloem, R., Ehlers, R., K{\"o}nighofer, B., Niekum, S., Topcu, U.:
  Safe reinforcement learning via shielding. In: Thirty-Second AAAI Conference
  on Artificial Intelligence (2018)

\bibitem{alur:mtl}
Alur, R., Henzinger, T.A.: Logics and models of real time: A survey. In:
  de~Bakker, J.W., Huizing, C., de~Roever, W.P., Rozenberg, G. (eds.)
  Real-Time: Theory in Practice. pp. 74--106. Springer Berlin Heidelberg (1992)

\bibitem{annpureddyLFS11}
Annpureddy, Y., Liu, C., Fainekos, G.E., Sankaranarayanan, S.: S-taliro: {A}
  tool for temporal logic falsification for hybrid systems. In: Tools and
  Algorithms for the Construction and Analysis of Systems, {TACAS}. pp.
  254--257 (2011)

\bibitem{brockman2016openai}
Brockman, G., Cheung, V., Pettersson, L., Schneider, J., Schulman, J., Tang,
  J., Zaremba, W.: Openai gym. arXiv preprint arXiv:1606.01540  (2016)

\bibitem{donze2009parameter}
Donz{\'e}, A., Krogh, B., Rajhans, A.: Parameter synthesis for hybrid systems
  with an application to {Simulink} models. In: International Workshop on
  Hybrid Systems: Computation and Control. pp. 165--179. Springer (2009)

\bibitem{dreossi-nfm17}
Dreossi, T., Donze, A., Seshia, S.A.: Compositional falsification of
  cyber-physical systems with machine learning components. In: Proceedings of
  the NASA Formal Methods Conference (NFM). pp. 357--372 (May 2017)

\bibitem{dreossi-ijcai18}
Dreossi, T., Ghosh, S., Yue, X., Keutzer, K., Sangiovanni-Vincentelli, A.,
  Seshia, S.A.: Counterexample-guided data augmentation. In: 27th International
  Joint Conference on Artificial Intelligence (IJCAI). pp. 2071--2078 (2018)

\bibitem{dreossi-cav18}
Dreossi, T., Jha, S., Seshia, S.A.: Semantic adversarial deep learning. In:
  30th International Conference on Computer Aided Verification (CAV) (2018)

\bibitem{duggirala2015c2e2}
Duggirala, P.S., Mitra, S., Viswanathan, M., Potok, M.: {C2E2}: a verification
  tool for stateflow models. In: International Conference on Tools and
  Algorithms for the Construction and Analysis of Systems. pp. 68--82. Springer
  (2015)

\bibitem{scenic}
Fremont, D.J., Yue, X., Dreossi, T., Ghosh, S., Sangiovanni-Vincentelli, A.L.,
  Seshia, S.A.: Scenic: Language-based scene generation. CoRR
  \textbf{arXiv:1809.09310} (2018), \url{http://arxiv.org/abs/1809.09310}

\bibitem{gehr2018ai}
Gehr, T., Mirman, M., Drachsler-Cohen, D., Tsankov, P., Chaudhuri, S., Vechev,
  M.: {AI2}: Safety and robustness certification of neural networks with
  abstract interpretation. In: Security and Privacy (SP), 2018 IEEE Symposium
  on (2018)

\bibitem{godefroid2008grammar}
Godefroid, P., Kiezun, A., Levin, M.Y.: Grammar-based whitebox fuzzing. In: ACM
  SIGPLAN Notices. vol.~43, pp. 206--215. ACM (2008)

\bibitem{guideways}
Grembek, O., Kurzhanskiy, A.A., Medury, A., Varaiya, P., Yu, M.: Making
  intersections safer with {I2V} communication (2019),
  \url{http://arxiv.org/abs/1803.00471}, to appear in Transportation Research,
  Part C.

\bibitem{halton1960efficiency}
Halton, J.H.: On the efficiency of certain quasi-random sequences of points in
  evaluating multi-dimensional integrals. Numerische Mathematik  \textbf{2}(1),
   84--90 (1960)

\bibitem{SchulmanWDRK17}
Schulman, J., Wolski, F., Dhariwal, P., Radford, A., Klimov, O.: Proximal
  policy optimization algorithms. CoRR  \textbf{abs/1707.06347} (2017),
  \url{http://arxiv.org/abs/1707.06347}

\bibitem{seshia-arxiv16}
Seshia, S.A., Sadigh, D., Sastry, S.S.: {Towards Verified Artificial
  Intelligence}. ArXiv e-prints  (July 2016)

\bibitem{tuncali-ivs18}
Tuncali, C.E., Fainekos, G., Ito, H., Kapinski, J.: Simulation-based
  adversarial test generation for autonomous vehicles with machine learning
  components. In: 2018 {IEEE} Intelligent Vehicles Symposium, {IV} 2018,
  Changshu, Suzhou, China, June 26-30, 2018. pp. 1555--1562 (2018)

\bibitem{pymtl}
Vazquez-Chanlatte, M.: mvcisback/py-metric-temporal-logic: v0.1.1 (Jan 2019).
  \doi{10.5281/zenodo.2548862}, \url{https://doi.org/10.5281/zenodo.2548862}

\bibitem{Webots}
Webots: http://www.cyberbotics.com, \url{http://www.cyberbotics.com},
  commercial Mobile Robot Simulation Software

\bibitem{wicker2018feature}
Wicker, M., Huang, X., Kwiatkowska, M.: Feature-guided black-box safety testing
  of deep neural networks. In: International Conference on Tools and Algorithms
  for the Construction and Analysis of Systems. pp. 408--426. Springer (2018)

\bibitem{squeezedet}
Wu, B., Iandola, F., Jin, P.H., Keutzer, K.: Squeezedet: Unified, small, low
  power fully convolutional neural networks for real-time object detection for
  autonomous driving (2016)

\end{thebibliography}
%
%
%
%
%
\end{document}